%% file: main.tex
\documentclass{article}
\usepackage{bbm}
\usepackage[table]{xcolor}
\usepackage{multirow}
\usepackage[english]{babel}

\usepackage[letterpaper,top=2cm,bottom=2cm,left=3cm,right=3cm,marginparwidth=1.75cm]{geometry}

\usepackage{amsmath}
\usepackage{graphicx}
\usepackage[colorlinks=true, allcolors=blue]{hyperref}
\usepackage{booktabs} 
\usepackage{caption}
\input{macro}

\title{FrugalGPT: How to Use Large Language Models  \\While Reducing Cost and Improving Performance}
\author{Lingjiao Chen, Matei Zaharia, James Zou\\\\
Stanford University}
\date{}

\begin{document}
\maketitle

\begin{abstract}
There is a rapidly growing number of large language models (LLMs) that users can query for a fee. We review the cost associated with querying popular LLM APIs---e.g. GPT-4, ChatGPT, J1-Jumbo---and find that these models have heterogeneous pricing structures, with fees that can differ by two orders of magnitude. In particular, using LLMs on large collections of queries and text can be expensive. Motivated by this, we outline and discuss  three types of strategies that users can exploit to reduce the inference cost associated with using LLMs: 1) prompt adaptation, 2) LLM approximation, and 3) LLM cascade. As an example, we propose \systemFGPT{}, a simple yet flexible instantiation of LLM cascade which learns which combinations of LLMs to use for different queries in order to reduce cost and improve accuracy. Our experiments show that \systemFGPT{} can match the performance of the best individual LLM (e.g. GPT-4) with up to 98\% cost reduction or improve the accuracy over GPT-4 by 4\% with the same cost. The ideas and findings presented here lay a foundation for using LLMs sustainably and efficiently.     

\end{abstract}

\section{Introduction}

\begin{figure}
\centering
\includegraphics[width=0.99\textwidth]{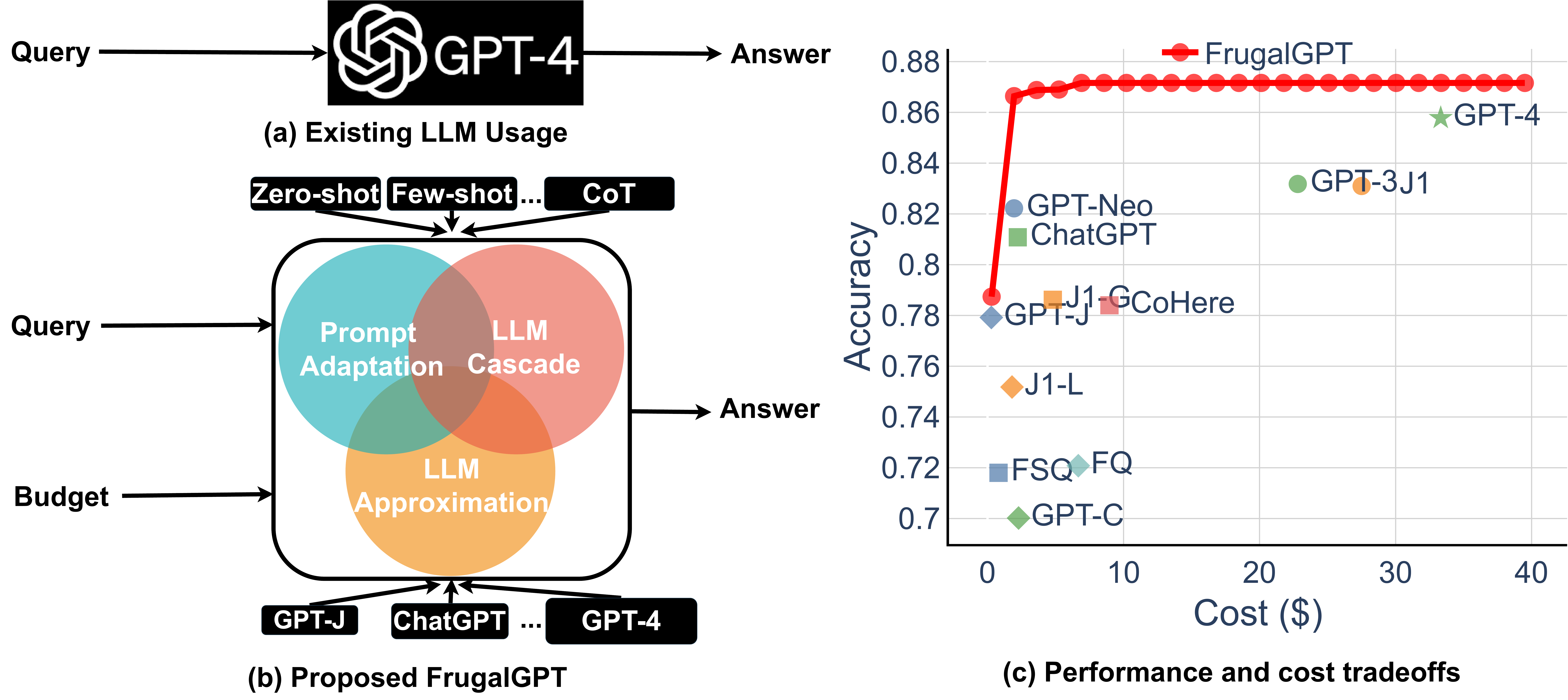}
\caption{\label{fig:AML:Intro} 
Our vision for reducing LLM cost while improving accuracy. (a) The standard usage sends queries to a single LLM (e.g. GPT-4), which can be expensive. (b) Our proposal is to use prompt adaption, LLM approximation and LLM cascade to reduce the inference cost. By optimizing over the selection of  different LLM APIs (e.g., GPT-J, ChatGPT, and GPT-4) as well as prompting strategies (such as zero-shot~\cite{brown2020language}, few-shot~\cite{liu2021makes}, and chain-of-thought(CoT)~\cite{wei2022chain}), we can achieve substantial efficiency gains. 
(c) On HEADLINES (a financial news dataset),  \systemFGPT{} can reduce the inference cost by 98\% while exceeding the performance of the best individual LLM (GPT-4).} 
\end{figure}

We are in the midst of an explosion of large language models (LLMs). The alluring possibilities of using LLMs for large-scale applications such as commerce, science, and finance have led a growing number of companies (OpenAI, AI21, CoHere, etc.) to offer LLMs as services.  


While LLMs such as GPT-4 achieves unprecedented performance in tasks such as question answering, using them for high-throughput applications can be very expensive. For example, ChatGPT is estimated to cost over \$700,000 per day to operate~\cite{CostEstChatGPT}, and using GPT-4 to support customer service can cost a small business over \$21,000 a month~\cite{CostEstGPT3}. In addition to the financial cost, using the largest LLMs encures substantial environmental and energy impact~\cite{bender2021dangers,wu2022sustainable}, affecting the social welfare of current and future generations. 

There are many LLMs now available via APIs and they charge heterogeneous prices. 
The cost of using a LLM API typically 
 consists of three components: 1) prompt cost (proportional to the length of the prompt), 2) generation cost (proportional to the generation length), and 3) sometimes a fixed cost per query. We  compared the cost associated with using 12 different commercial LLMs from mainstream providers including OpenAI, AI21, CoHere and Textsynth (Table \ref{tab:AML:LLMAPIs}). Their cost can differ by up to 2 orders of magnitudes: for example,  the prompt cost for 10M tokens is \$30 for OpenAI's GPT-4 but only \$0.2 for GPT-J hosted by Textsyth.

Given the heterogeneous cost and quality, how to effectively and efficiently leverage the full set of LLM options is a key challenge for pracitioners. 
If the tasks are relatively simple, then aggregating multiple responses from GPT-J~\cite{wang2021gpt} (whose size is 30x smaller than GPT-3) offers performance similar to GPT-3~\cite{arora2022ask}, leading to financial and environmental savings.
However, the  performance of GPT-J can be much worse on difficult tasks~\cite{touvron2023llama}. Moreover, relying on one API provider is not reliable if that provider becomes unavailable, potentially due to spiking demand.
Existing model ensemble paradigms such as model cascade~\cite{viola2004robust,wang2011cascade} and  FrugalML~\cite{chen2020frugalml,chen2022efficient} were designed for predictive tasks with a known set of labels and do not account for the full capabilities of LLM.
How to use LLMs affordably and accurately therefore calls for new approaches.

\paragraph{Our contributions.} In this paper, we lay out our vision of a flexible framework that uses LLM APIs to process natural language queries within a  budget constraint, termed \systemFGPT{}. 
As shown in Figure \ref{fig:AML:Intro}, we discuss three main strategies for cost reduction: \textit{prompt adaptation},    \textit{LLM approximation},  
and \textit{LLM cascade}.
The prompt adaptation explores how to identify effective (often shorter)  prompts to save cost. 
LLM approximation aims to create simpler and cheaper LLMs to match a powerful yet expensive LLM on  specific  tasks.  
LLM cascade focuses on how to adaptively choose which LLM APIs to use for different queries. 

To illustrate the potential of these ideas, we  implement and evaluate a simple version of \systemFGPT{} using LLM cascade. On each dataset and task, \systemFGPT{} learns how to adaptively triage different queries in the dataset to different combinations of LLMs, including ChatGPT~\cite{ChatGPTAnn}, GPT-3~\cite{brown2020language} and GPT-4~\cite{GPT4TechReport}. Our experiments show that \systemFGPT{} can save up to 98\% of the inference cost of the best individual LLM API while matching its  performance on the downstream task.
On the other hand, \systemFGPT{} can improve the performance by up to 4\% with the same cost.
We believe this is only the tip of the iceberg and we hope \systemFGPT{} opens a new window toward reducing LLMs' inference cost and improving its performances.

\paragraph{Related Works.} \textbf{Prompt Engineering.} Prompt engineering has emerged as a discipline for crafting prompts to enhance LLMs' performance across various applications. Recent developments include few-shot~\cite{brown2020language}, chain-of-thought~\cite{wei2022chain}, knowledge enhancement~\cite{liu2021generated,khattab2022demonstrate}, and numerous other prompting techniques~\cite{mialon2023augmented,khot2022decomposed,zhou2022least,dua2022successive}. Existing prompt engineering approaches often aim to provide more detailed task explanations and in-context examples, resulting in longer and more expensive prompts. In contrast, this paper explores the use of concise prompts to reduce costs.

\textbf{Model Ensemble.} Model ensembles, which involve combining multiple ML models for prediction, have gained popularity in supervised learning~\cite{viola2004robust, friedman2002stochastic}, unsupervised learning~\cite{yang2014exploring}, semi-supervised learning~\cite{gupta2022semi}, and weakly supervised learning~\cite{diba2017weakly}. Model ensembles typically require white-box access to multiple models for training purposes, but LLM APIs are often black-box. Moreover, model ensembles necessitate querying all models for a single query, thereby increasing costs.

\textbf{System Optimization for LLMs.} Numerous efforts have aimed to accelerate the training and inference time of modern deep learning models through system optimization~\cite{han2015deep,cai2017deep,cass2019taking,jia2019beyond,recht2011hogwild}. Recent work focuses on post-training quantization~\cite{bai2022towards, yao2023comprehensive,xiao2022smoothquant}, training pipeline parallelism~\cite{li2021terapipe}, and hardware-aware pruning~\cite{kurtic2023ziplm} tailored for LLMs. System optimization requires modifications to LLMs' internal states (e.g., model weights), but many commercial LLM APIs do not release their models. Additionally, the rapidly increasing size of LLMs renders retraining highly expensive.

\textbf{ML-as-a-Service.} LLM APIs constitute a crucial component of the rapidly expanding machine-learning-as-a-service (MLaaS) industry. Recent studies have demonstrated the diversity of different ML APIs' predictions~\cite{buolamwini2018gender,koenecke2020racial,chen2021did} and proposed strategies for leveraging various classification ML APIs to improve performance~\cite{chen2020frugalml,chen2022efficient}. The outputs of LLM APIs encompass the entire natural language space, but existing work requires a fixed (and known) label set. Moreover, both prompt choices and LLM API selections significantly impact generative tasks' performance, resulting in a considerably larger optimization space than standard classification.

The remaining part of the paper is organized as follows.
We start by offering more context and the problem statement in Section \ref{sec:AML:Prelim}.
Next in Section \ref{sec:AML:techniques}, we present our visions on how to use LLM  APIs affordability and accurately.
Section \ref{sec:AML:Experiment} shows the empirical benefits of \systemFGPT{}  using real-world LLM APIs (including GPT-3, ChatGPT, and GPT-4). Finally, we discuss future prospects in Section \ref{sec:AML:conclusion}.

\section{Scope and Problem Statement}\label{sec:AML:Prelim}

\paragraph{Natural language query answering.}
In this paper, we concentrate on the standard natural language query answering task, where the objective is to answer a query $\query$ sampled from a natural language query distribution $\mathcal{Q}$. Various real-world natural language tasks, such as news classification, reading comprehension, and commonsense reasoning,  can be formulated as query-answering problems.

\paragraph{LLM marketplace.}
We consider answering queries via the LLM market, which comprises $K$ different LLM APIs, denoted by $\{f_i(\cdot)\}_{i=1}^{K}$. Each $f_i(\cdot): \mathcal{P}\mapsto \mathcal{A}$ is a function that, given a prompt $\prompt$ from the prompt space $\mathcal{P}$, generates an answer  from the answer distribution $\mathcal{A}$. 
Note that to use LLM APIs, one has to convert each query $\query$ to some corresponding prompt first. LLM APIs are associated with their own \textit{cost}, typically consisting of three components: a portion proportional to the length of the prompt, a portion proportional to the length of the generated answer, and (sometimes) a fixed cost per query. Formally, given a prompt $p$, the cost of using the $i$th LLM API is denoted by $\cost_i(\prompt) \triangleq \tilde{c}_{i,2}\|f_i(\prompt)\|+ \tilde{c}_{i,1} \|\prompt\| + \tilde{c}_{i,0}$, where $\tilde{c}_{i,j},j=0,1,2$ are constants. 

\paragraph{An illustrative example.} Adapting the case study provided by \cite{CostEstChatGPT}, assume a small business operates a customer service using GPT-4. The company caters to 15,000 customers each month, with each customer asking three questions twice a week, totaling 360,000 queries per month. Suppose for each question, its corresponding prompt averages 1800 tokens, and the answer is around 80 tokens. Considering that the input and response costs of GPT-4 are \$0.03 and \$0.06 per thousand tokens, respectively, the total monthly cost amounts to $360\times (\$0.03\times 1800+\$0.06\times 80)\approx\$21.2K$. 
 Such a high cost is prohibitive for many small businesses.

\paragraph{Problem statement: budget-aware LLM API usage.} Our primary goal in this paper is \textit{leveraging LLM APIs within a budget constraint}. Formally, this can be formulated as maximizing the overall task performance $\E_{(q,a)\in\mathcal{Q}\times \mathcal{A}}[r(a,\hat{a}(s,q))]$, while ensuring the average cost is bounded by a user-defined value $b$, i.e., $\E_{(q,a)\in\mathcal{Q}\times \mathcal{A}}[\cost(s,q)]\leq b$. Here, $a$ denotes the correct answer to the query $q$, $\hat{a}(s,q)$ is the generated answer by some strategy $s$ for query $q$, and $c(s,q)$ is the associated cost for processing query $q$ using strategy $s$. The reward function $r(\cdot,\cdot)$ measures how closely the generated answer aligns with the correct one. It is crucial to note that the search space for the strategy is vast, encompassing factors such as which prompts to use, which LLM APIs to employ, and how to aggregate their responses.

\eat{Colloquially, an LLM is a pre-train artificial intelligence model that takes a text query and a text prompt as the inputs, and generates some text response as the output.
The text query is generic: text classification, reading comprehension, story completion, language translation, and almost all other natural language tasks can be formed as a text query. 
The text prompt serves as a hint of how to answer the query. 
LLM APIs provide users with easy access to LLMs without bothering with data, software, and hardware to develop and maintain users' own LLMs.

Yet, LLM APIs come with their own \textit{cost}. It typically consists of three parts: one portion proportional to the length of the query and prompt, one portion  proportional to the length of the generated answer, and sometimes a fixed cost per query.
}

\section{How to Use LLMs Affordably and Accurately}\label{sec:AML:techniques}
Now we present our vision on how to use LLM APIs within a budget.  
As shown in Figure \ref{fig:AML:Intro} (b), we discuss three cost-reduction strategies, namely, prompt adaptation,  LLM approximation,  and LLM cascade. 

\begin{figure}
\centering
\includegraphics[width=0.99\textwidth]{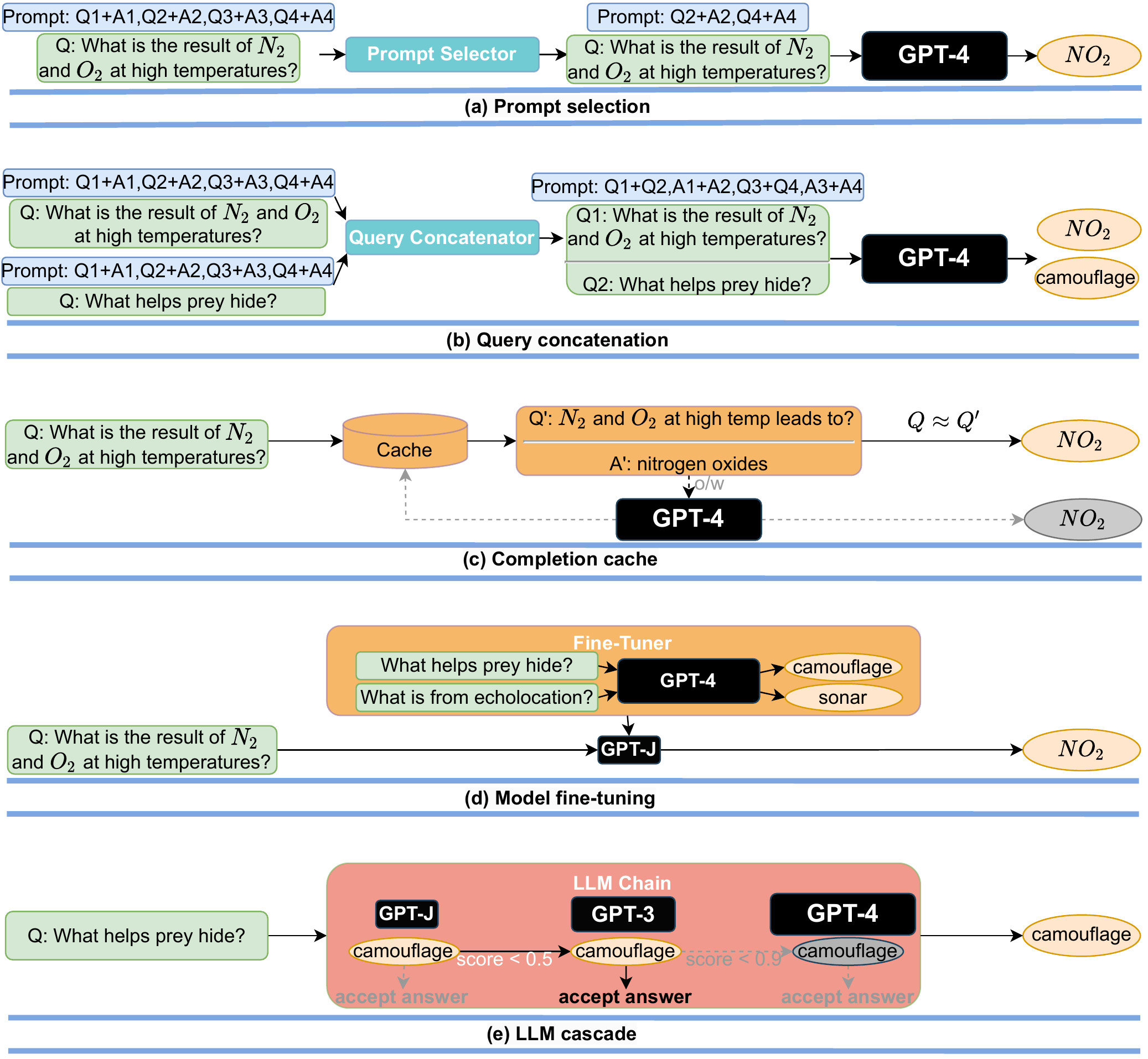}
\caption{\label{fig:AML:Examples}  Illustrations of cost-saving strategies. (a) Prompt selection uses a subset of in-context examples as the prompt to reduce the size of the prompt. (b) Query concatenation aggregates multiple queries to share prompts. (c) Completion cache stores and reuses an LLM API's response when a similar query is asked. (d) Model fine-tuning uses expensive LLMs' responses to fine-tune cheap LLMs. (e) LLM cascade employs different LLM APIs for different queries.  }
\end{figure}

\paragraph{Strategy 1: Prompt adaptation.} 
The cost of an LLM query increases linearly with the size of the prompt. Consequently, a logical approach to reduce the cost of using LLM APIs involves decreasing the prompt's size, a process we refer to as prompt adaptation. \textit{Prompt selection} (as illustrated in Figure \ref{fig:AML:Examples} (a)) is a natural example of prompt adaptation: rather than employing a prompt containing numerous examples that demonstrate how to perform a task, one can retain a small subset of examples in the prompt. This results in a smaller prompt and subsequently lower cost. An intriguing challenge of prompt selection lies in determining which examples to maintain for various queries without compromising task performance.

An additional instantiation is \textit{query concatenation} (Figure \ref{fig:AML:Examples} (b)). It is important to note that processing queries individually necessitates sending the same prompt to an LLM API multiple times. Therefore, the fundamental concept of query concatenation involves sending the prompt only once to the LLM API while allowing it to address multiple queries, thereby preventing redundant prompt processing. To accomplish this, several queries must be concatenated into a single query, and the prompt must explicitly request the LLM API to process multiple queries. For instance, to handle two queries using one prompt, the examples presented in the prompt can include both queries followed by their corresponding answers.

\paragraph{Strategy 2: LLM approximation.} The concept of \textit{LLM approximation} is quite simple: if an LLM API is too costly to utilize, one can approximate it using more affordable models or infrastructures. One example is the \textit{completion cache}: as depicted in Figure \ref{fig:AML:Examples} (c), the fundamental idea involves storing the response locally in a cache (e.g., a database) when submitting a query to an LLM API. To process a new query, we first verify if a similar query has been previously answered. If so, the response is retrieved from the cache. An LLM API is invoked only if no similar query is discovered in the cache. The completion cache provides substantial cost savings when similar queries are frequently posed. For instance, consider a search engine powered by an LLM API. If numerous users search for the same or similar keywords simultaneously, the completion cache facilitates answering all their queries by invoking the LLM only once.

Another example of LLM approximation is \textit{model fine-tuning}. As shown in Figure \ref{fig:AML:Examples}(d), this process consists of three steps: first, collect a powerful but expensive LLM API's responses to a few queries; second, use the responses to fine-tune a smaller and more affordable AI model; and finally, employ the fine-tuned model for new queries. In addition to cost savings, the fine-tuned model often does not require lengthy prompts, thus providing latency improvements as a byproduct.

\paragraph{Strategy 3: LLM cascade.}
The increasing availability of LLM APIs with heterogeneous performance and costs presents a unique opportunity for data-adaptive LLM selection. Different LLM APIs have their own strengths and weaknesses for various queries. Consequently, appropriately selecting which LLMs to use can provide both cost reduction and performance improvements. \textit{LLM cascade}, as illustrated in Figure \ref{fig:AML:Examples} (e), is one such example. \textit{LLM cascade} sends a query to a list of LLM APIs sequentially. If one LLM API's response is reliable, then its response is returned, and no further LLMs in the list are needed. The remaining LLM APIs are queried only if the previous APIs' generations are deemed insufficiently reliable. Query cost is significantly reduced if the first few APIs are relatively inexpensive and produce reliable generations.

The key components of LLM cascade consist of two elements: (i) a generation scoring function and (ii) an LLM router. The generation scoring function, denoted by $g(\cdot,\cdot):\mathcal{Q} \times \mathcal{A}\mapsto [0,1]$, generates a reliability score given a query and an answer produced by an LLM API. The LLM router selects $m$ LLM APIs to include in the list. Let $\pmb L \in [K]^m$ denote the indexes of the $m$ APIs selected by the router. Given a new query, it iteratively invokes the $i$th API in the list to obtain an answer $f_{L_i}(\query)$. Then, it uses the scoring function to generate a score $ g(\query,f_{L_i}(\query))$. It returns the generation if the score is higher than a threshold $\pmb \tau_i$, and queries the next service otherwise.

The scoring function can be obtained by training a simple regression model that learns whether a generation is correct from the query and a generated answer. Learning the selected list $\pmb L$ and the threshold vectors $\pmb \tau$ can be modeled as a constraint optimization problem:
\begin{equation*}
    \begin{split}
        \max_{\pmb L, \pmb \tau}  \textit{ }& \E\left[r(a,f_{L_z}(q))\right]\\
    s.t. \textit{ }& \E\left[\sum_{i=1}^{z} \tilde{c}_{L_i,2} \|f_{L_i}(\query)\|+\tilde{c}_{L_i,1} \|\query\|+ \tilde{c}_{L_i,0}\right]\leq b,    \\
    & z = \arg\min_i g(q,f_{L_i}(q)) \geq  \pmb \tau_i
    \end{split}
\end{equation*}
Here, $z$ denotes the LLM API at which the router stops and returns the answer, the first constraint ensures the average cost is bounded by the budget, and the objective measures the quality of the generation $f_{L_z}(q)$ for a query $\query$ compared to the true answer $\answer$. This problem is inherently a mixed-integer optimization and thus computationally expensive to solve. To address this issue, we develop a specialized optimizer that (i) prunes the search space of $\pmb L$ by ignoring any list of LLMs with small answer disagreement, and (ii) approximates the objective by interpolating it within a few samples. This results in an efficient implementation with satisfactory performance, as  shown later in Figure \ref{fig:AML:Tradeoffs}.

\eat{Many LLM queries aim at picking one of many options. 
For example, information retrieval queries need an LLM API to select one from many documents that best answers a given question.
One popular approach~\cite{liang2022holistic} is to send the question and each possible document to the LLM API, and select the document on which an LLM API favors the most, but this is costly to process plenty of documents.
 \textit{Option pruning} helps mitigate this cost.
 As shown in Figure \ref{fig:AML:Examples} (f), the key idea is to use cheap LLMs to filter out documents that are highly likely to be irrelative, and then only use the expensive LLM to process a few options. As long as the cheap LLM APIs keep the correct option in the filtered documents,  the performance will be at least as good as using the expensive API for all documents. 
}

\paragraph{Compositions.} Combining approaches within and across different strategies can lead to further cost reduction and performance enhancement. For instance, \textit{joint prompt and LLM selection} is a composition of prompt selection and LLM cascade: for a given query, it searches for the smallest prompt and most affordable LLM that achieves satisfactory task performance. Another example is to search across both existing LLM APIs and fine-tuned models. It is important to note that the composition of different approaches also increases the computational costs for training. Consequently, this paves the way for investigating trade-offs between query costs, task performance, and computational costs.

\section{LLM Cascade Reduces Cost and Improves Accuracy}\label{sec:AML:Experiment}
In this section, we present an empirical study on the \systemFGPT{} LLM cascade. Our goals are three-fold: (i) understand what a simple instantiation of LLM cascade learns, (ii) quantify the cost savings attained by \systemFGPT{} while matching the best individual LLM API's performance, and (iii) measure the trade-offs between performance and cost enabled by \systemFGPT{}.

\paragraph{Setups: LLM APIs, Tasks, Datasets, and \systemFGPT{} instances.}
We have selected 12 LLM APIs from 5 mainstream providers, namely, OpenAI~\cite{OpenAIAPI}, AI21~\cite{AI21API}, CoHere~\cite{CoHereAPI}, Textsynth~\cite{TextsynthAPI}, and ForeFrontAI~\cite{FFAIAPI}. The details are summarized in Table \ref{tab:AML:LLMAPIs}. \systemFGPT{} has been developed on top of these APIs and evaluated on a range of datasets belonging to different tasks, including HEADLINES~\cite{sinha2021impact}, OVERRULING~\cite{zheng2021does}, and COQA~\cite{reddy2019coqa}. The summary of these datasets is presented in Table \ref{tab:AML:Data}.
HEADLINES is a financial news dataset whose goal is to determine the gold price trend (up, down, neutral, or none) by reading financial news titles. This is especially useful for filtering relevant news in financial markets. OVERRULING is a legal document dataset where the goal is to determine whether a given sentence is an overruling, i.e., rejecting previous legal cases. COQA is a reading comprehension dataset developed in a conversational setting, which we have adapted as a direct query answering task.
We focus on the LLM cascade approach with a cascade length of 3, as this simplifies the optimization space and already demonstrates good results. Each dataset is randomly split into a training set to learn the LLM cascade and a test set for evaluation. 

\begin{table}[htbp]
  \centering
  \small
  \caption{Summary of commercial LLM APIs. We use 12 LLM APIs from 5 providers. The cost was retrieved in March 2023. The cost can have three additive components: input (proportional to the number of input tokens), output (proportional to the number of generated tokens) and a fixed cost per request. The LLMs's costs can differ by up to 2 orders of magnitudes. For example, to process 10M input tokens, GPT-J from Textsynth costs only \$0.2, but OpenAI's GPT-4 needs \$30.}
    \begin{tabular}{|c|c|c|c|c|c|}
    \hline
    \multirow{2}[4]{*}{\textbf{Provider}} &
      \multirow{2}[4]{*}{\textbf{API}} &
      \multicolumn{1}{c|}{\multirow{2}[4]{*}{\textbf{Size/B}}} &
      \multicolumn{3}{c|}{\textbf{Cost (USD)}}
      \bigstrut\\
\cline{4-6}    \multicolumn{1}{|c|}{} &
      \multicolumn{1}{c|}{} &
       &
      \multicolumn{1}{c|}{\textbf{10M input tokens}} &
      \multicolumn{1}{c|}{\textbf{10M output tokens}} &
      \multicolumn{1}{c|}{\textbf{ request}}
      \bigstrut\\
    \hline
    \hline
    \multirow{4}[8]{*}{\textbf{OpenAI}} &
      GPT-Curie &
      6.7 &
      2 &
      2 &
      0
      \bigstrut\\
\cline{2-6}    \multicolumn{1}{|c|}{} &
      ChatGPT &
      \multicolumn{1}{c|}{NA} &
      2 &
      2 &
      0
      \bigstrut\\
\cline{2-6}    \multicolumn{1}{|c|}{} &
      GPT-3 &
      175 &
      20 &
      20 &
      0
      \bigstrut\\
\cline{2-6}    \multicolumn{1}{|c|}{} &
      GPT-4 &
      \multicolumn{1}{c|}{NA} &
      30 &
      60 &
      0
      \bigstrut\\
    \hline
    \hline
    \multirow{3}[6]{*}{\textbf{AI21}} &
      J1-Large &
      7.5 &
      0 &
      30 &
      0.0003
      \bigstrut\\
\cline{2-6}    \multicolumn{1}{|c|}{} &
      J1-Grande &
      17 &
      0 &
      80 &
      0.0008
      \bigstrut\\
\cline{2-6}    \multicolumn{1}{|c|}{} &
      J1-Jumbo &
      178 &
      0 &
      250 &
      0.005
      \bigstrut\\
    \hline
    \hline
    \textbf{Cohere} &
      Xlarge &
      52 &
      10 &
      10 &
      0
      \bigstrut\\
    \hline      
    \hline
    \textbf{ForeFrontAI} &
      QA &
      16 &
      5.8 &
      5.8 &
      0
      \bigstrut\\
    \hline      
    \hline
    \multirow{3}[6]{*}{\textbf{Textsynth}} &
      GPT-J &
      6 &
      0.2 &
      5 &
      0
      \bigstrut\\
\cline{2-6}    \multicolumn{1}{|c|}{} &
      FAIRSEQ &
      13 &
      0.6 &
      15 &
      0
      \bigstrut\\
\cline{2-6}    \multicolumn{1}{|c|}{} &
      GPT-Neox &
      20 &
      1.4 &
      35 &
      0
      \bigstrut\\
    \hline
    \end{tabular}%
  \label{tab:AML:LLMAPIs}%
\end{table}%

\begin{table}[htbp]
  \centering
  \small
  \caption{Summary of datasets used in the \systemFGPT{} LLM cascade experiments.}
    \begin{tabular}{|c||c|c|c|}
    \hline
    Dataset & Domain & Size  & \#Examples in the prompt \bigstrut\\
    \hline
    \hline
    HEADLINES & Finance & 10000 & 8 \bigstrut\\
    \hline
    OVERRULING & Law   & 2400  & 5 \bigstrut\\
    \hline
    COQA  & Passage Reading & 7982  & 2 \bigstrut\\
    \hline
    \end{tabular}%
  \label{tab:AML:Data}%
\end{table}%

\paragraph{A Case Study.} 
\begin{figure}
\centering
\includegraphics[width=0.99\textwidth]{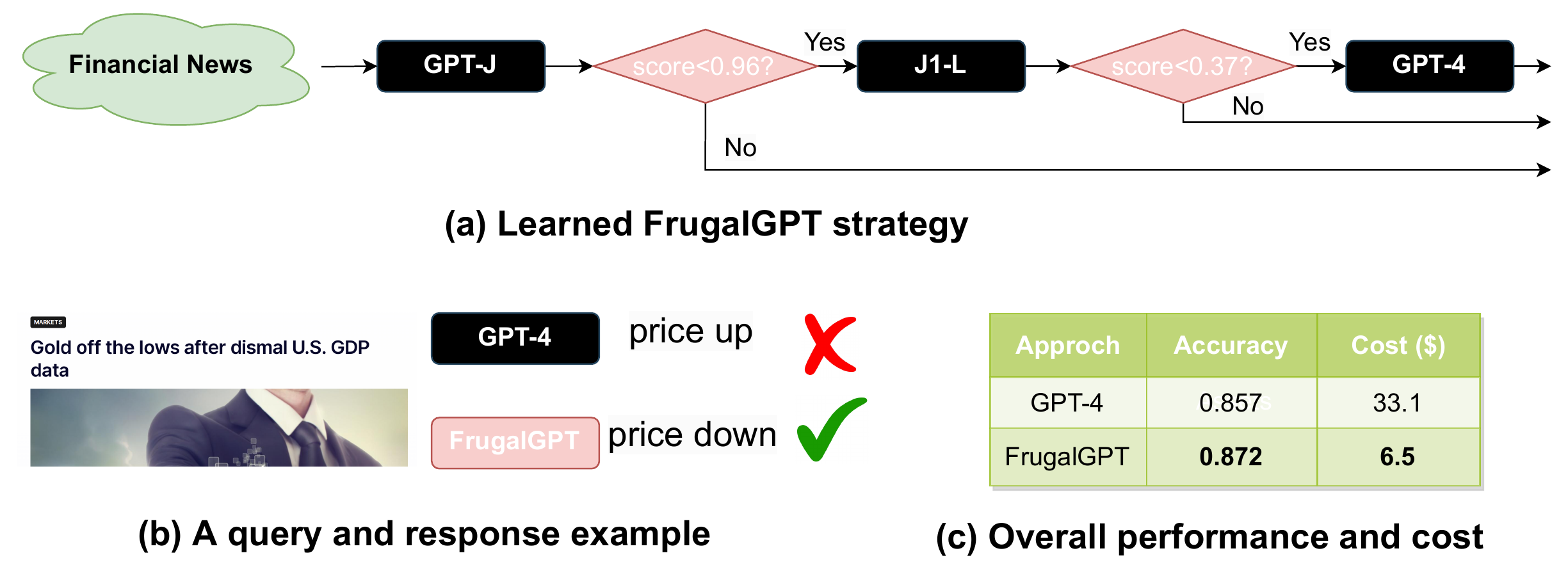}
\caption{A case study of \systemFGPT{} on the HEADLINES dataset. (a) The cascade strategy that \systemFGPT{} learned on this dataset with overall budget \$6.5, one fifth of GPT-4's cost. \systemFGPT{} avoids querying GPT-4 as long as GPT-J and J1-L produce high-quality answers. (b) Sometimes GPT-4 makes a mistake, but \systemFGPT{}  learns to use the correct answers by J-1 and GPT-J.  
(c) Overall, we observe that \systemFGPT{} reduces the cost by 80\%, while improves the accuracy by 1.5\% compared to GPT-4.}\label{fig:AML:CaseStudy}
\end{figure}

Let us begin with a case study on the HEADLINES dataset. We set the budget to be \$6.5, which is one-fifth of GPT-4's cost. We employ a DistilBERT~\cite{sanh2019distilbert} tailored to regression as the scoring function. It is important to note that DistilBERT is considerably smaller and therefore less expensive than all LLMs considered here. As depicted in Figure \ref{fig:AML:CaseStudy} (a), the learned \systemFGPT{} sequentially calls GPT-J, J1-L, and GPT-4. For any given query, it first extracts an answer from GPT-J. If the score of this answer is greater than 0.96, the answer is accepted as the final response. Otherwise, J1-L is queried. J1-L's answer is accepted as the final response if its score is greater than 0.37; otherwise, GPT-4 is invoked to obtain the final answer. Interestingly, this approach outperforms GPT-4 for numerous queries. For instance, given a headline "Gold off the lows after dismal U.S. GDP data" from NASDAQ, \systemFGPT{} accurately predicts that the price is going down, while GPT-4 provides an incorrect answer (as shown in Figure \ref{fig:AML:CaseStudy}(b)). Overall, \systemFGPT{} results in both accuracy gains and cost reduction. As illustrated in Figure \ref{fig:AML:CaseStudy}(c), its cost is reduced by 80\%, while the accuracy is even 1.5\% higher.

\paragraph{LLM diversity.} Why can multiple LLM APIs potentially produce better performance than the best individual LLM? In essence, this is due to generation diversity: even an inexpensive LLM can sometimes correctly answer queries on which a more expensive LLM fails. To measure this diversity, we use the maximum performance improvement, or \textit{MPI}. The MPI of LLM A with respect to LLM B is the probability that LLM A generates the correct answer while LLM B provides incorrect ones. This metric essentially measures the maximum performance gains achievable by invoking LLM A in addition to LLM B.

\begin{figure}[t!]
     \centering
\begin{subfigure}[b]{0.328\textwidth}
         \centering
\includegraphics[width=\textwidth]{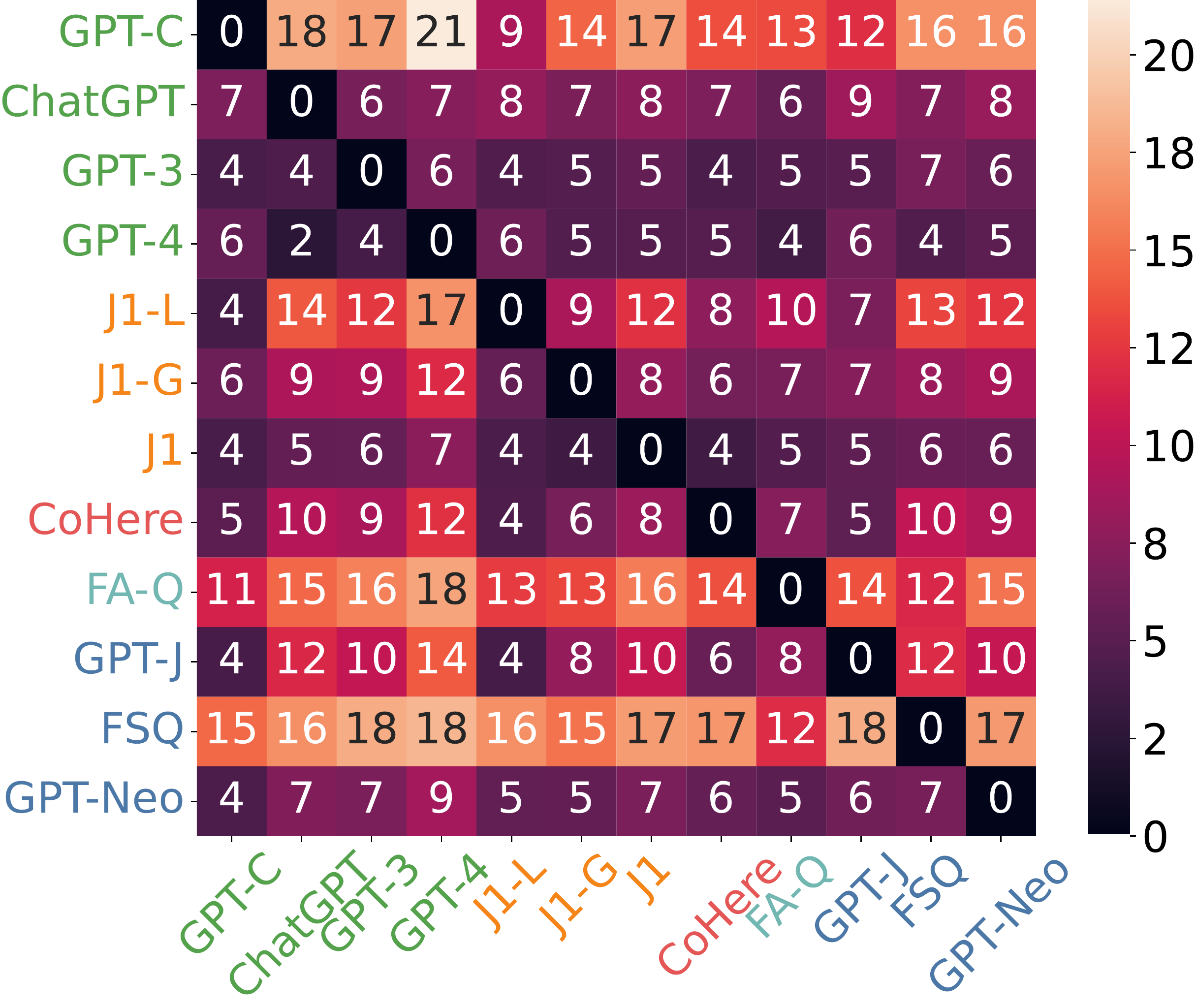}
         \caption{HEADLINES}
         \label{}
     \end{subfigure}
\begin{subfigure}[b]{0.328\textwidth}
         \centering
\includegraphics[width=\textwidth]{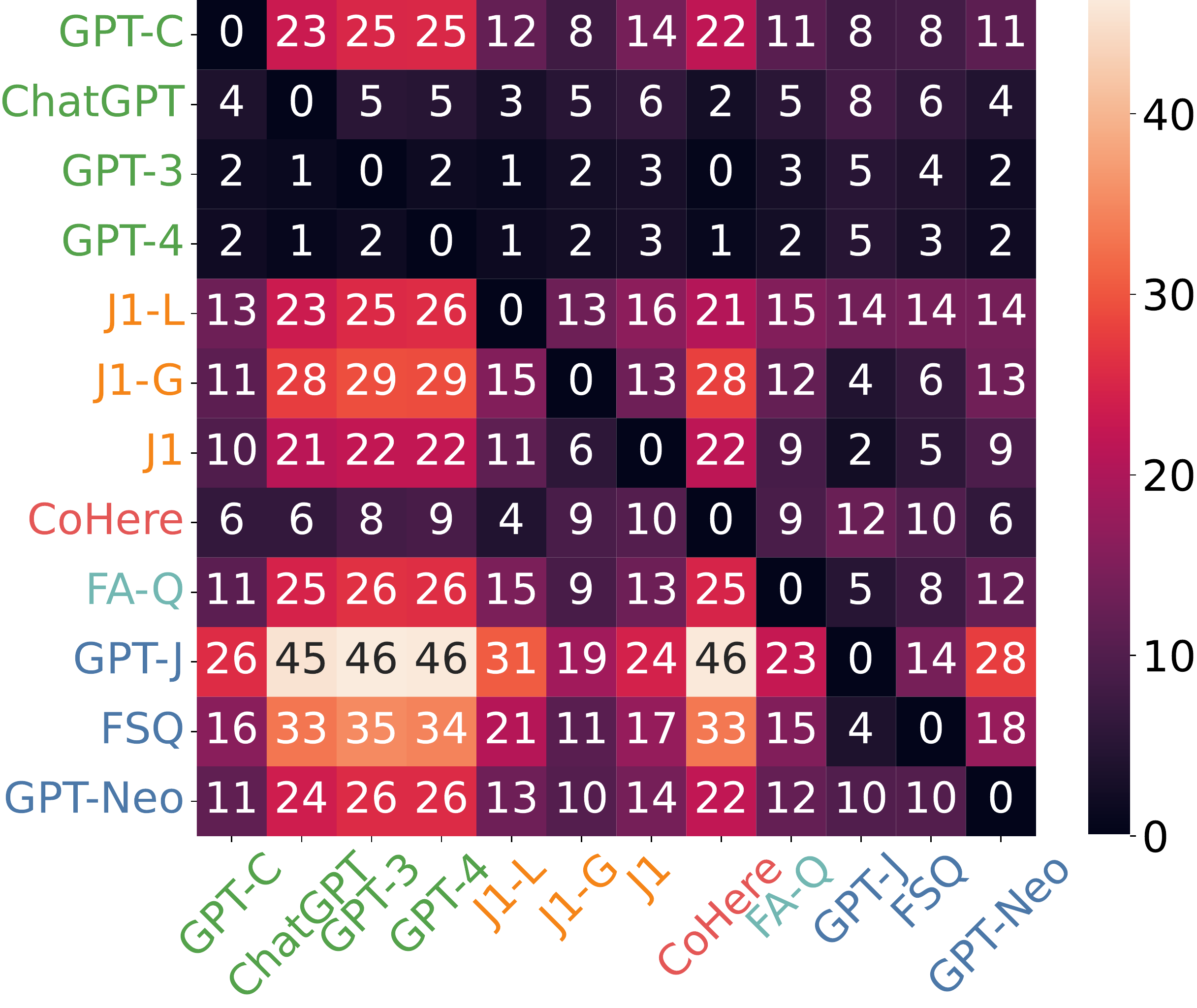}
         \caption{OVERRULING}
         \label{}
     \end{subfigure}
\begin{subfigure}[b]{0.328\textwidth}
         \centering
\includegraphics[width=\textwidth]{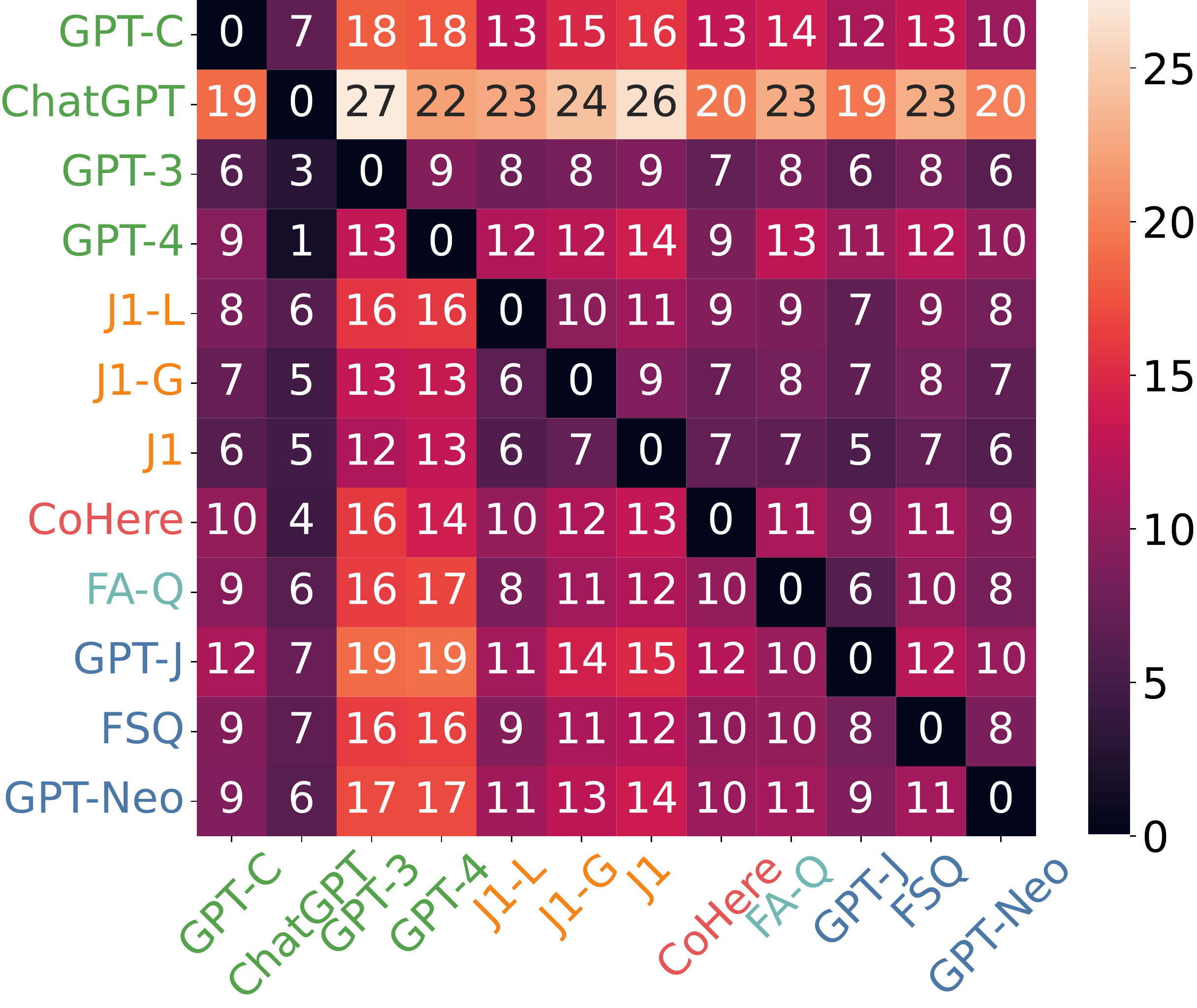}
         \caption{COQA}
         \label{}
     \end{subfigure}
     
        \caption{Maximum performance improvement (MPI) of each pair of LLMs. (a), (b), and (c) correspond to the three datasets, separately. One entry indicates the percent of cases that the LLM on its row is wrong but the LLM on its column gives the right answer.  Overall, we observe that cheap LLMs can be complementary to the expensive ones quite often. For example, for about 6\% of the data, GPT-4 makes a mistake but GPJ-J (or J-L or GPT-C) gives the right answer on HEADLINES. }
\label{fig:AML:MaxGap}
\end{figure}

MPI between each pair of LLM APIs for all datasets is displayed in Figure \ref{fig:AML:MaxGap}. Overall, we observe significant potential within the LLM marketplace. For instance, GPT-C, GPT-J, and J1-L can all enhance GPT-4's performance by up to 6\% on the HEADLINES dataset. On the COQA dataset, there are 13\% of data points where GPT-4 makes an error, but GPT-3 provides the correct answer. Although these improvement upper bounds may not always be attainable, they do demonstrate the possibility of utilizing more affordable services to achieve better performance.

\paragraph{Cost Savings.}

\begin{table}[htbp]
  \centering
  \caption{Cost savings by \systemFGPT{} to match the best individual LLM's performance.
  }
    \begin{tabular}{|c||c|c|c|c|}
    \hline
    \multirow{2}[4]{*}{Dataset} & \multirow{2}[4]{*}{Best invidual LLM} & \multicolumn{2}{c|}{Cost to reach the same accuracy} & \multirow{2}[4]{*}{Cost Savings} \bigstrut\\
\cline{3-4}          &       & Best individual LLM & FrugalGPT &  \bigstrut\\
    \hline
    \hline
    HEADLINES & GPT-4 & 33.1  & 0.6   & 98.3\% \bigstrut\\
    \hline
    OVERULLING & GPT-4 & 9.7   & 2.6   & 73.3\% \bigstrut\\
    \hline
    COQA  & GPT-3 & 72.5  & 29.6  & 59.2\% \bigstrut\\
    \hline
    \end{tabular}%
    \label{tab:AML:CostSavings}%
\end{table}%

Subsequently, we examine whether \systemFGPT{} can reduce costs while maintaining accuracy and, if so, by how much. Table \ref{tab:AML:CostSavings} displays the overall cost savings of \systemFGPT{}, which range from 50\% to 98\%. This is feasible because \systemFGPT{} identifies the queries that can be accurately answered by smaller LLMs and, as a result, only invokes those cost-effective LLMs. Powerful but expensive LLMs, such as GPT-4, are utilized only for challenging queries detected by \systemFGPT{}.

 \begin{figure}
\centering
\includegraphics[width=0.99\textwidth]{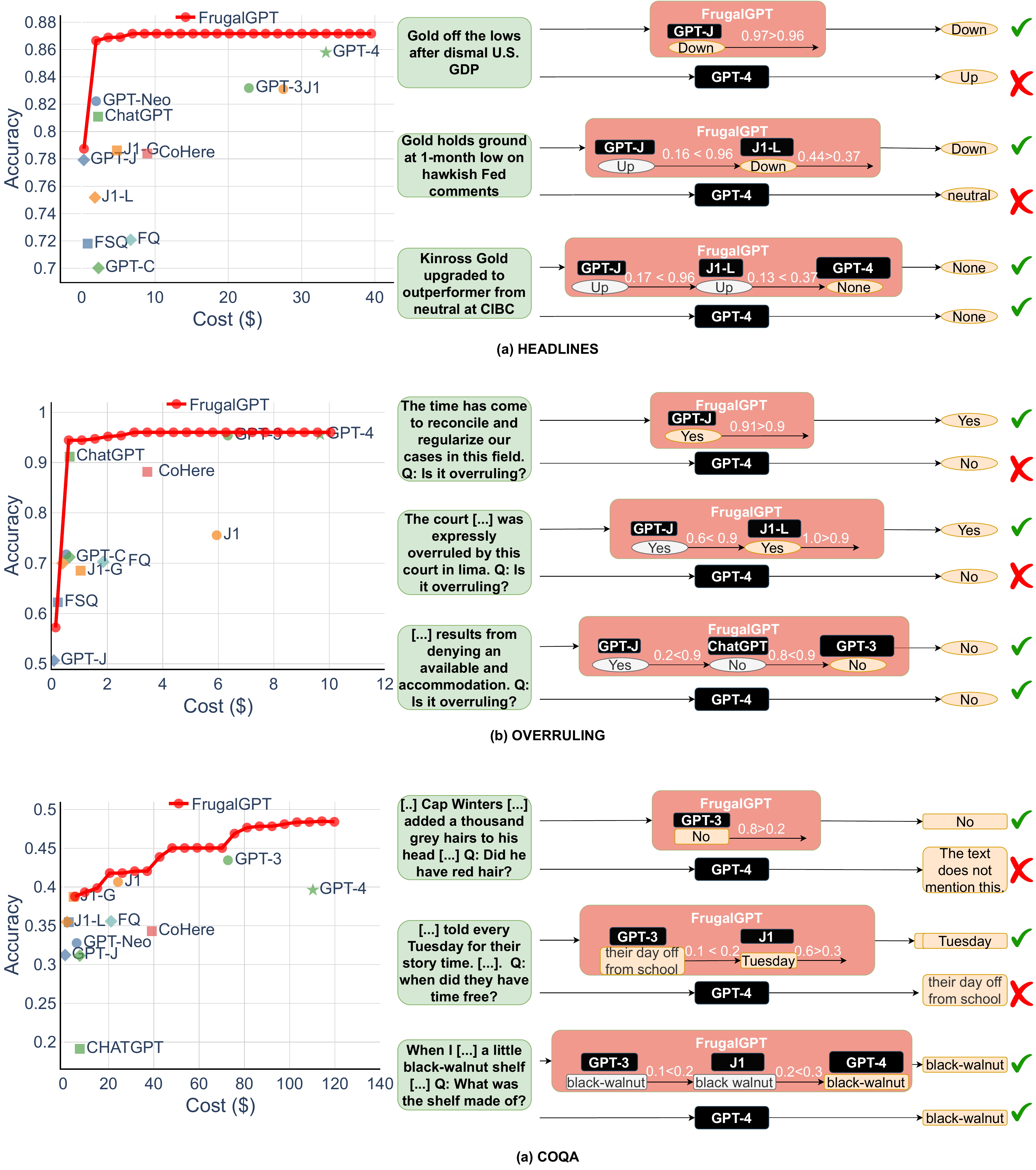}
\caption{Accuracy and cost tradeoffs achieved by \systemFGPT{}. Overall, \systemFGPT{} often achieves the same performance of the best individual LLM API (e.g., GPT-4) with orders of magnitudes smaller cost. 
        When incurring the same cost, \systemFGPT{} can improves the accuracy by up to 5\%. Examples of LLM cascade for each dataset are shown on the right. }\label{fig:AML:Tradeoffs}
\end{figure}
\paragraph{Performance and Cost Trade-offs.}
Now, we investigate the trade-offs between performance and cost achieved by \systemFGPT{}, as illustrated in Figure \ref{fig:AML:Tradeoffs}. Several interesting observations can be made. First, the cost ranking of different LLM APIs is not fixed. For instance, J1 is the second most expensive LLM on the HEADLINES dataset, while GPT-3 holds that position on the OVERRULING and COQA datasets. This is primarily due to the heterogeneous pricing mechanism: J1 incurs a high cost for each generated token but charges nothing for input tokens, whereas GPT-3 charges for both input and output tokens. Moreover, more expensive LLM APIs sometimes result in worse performance than their cheaper counterparts. For example, J1 is costlier than GPT-3 on HEADLINES, but its performance is inferior. These observations underscore the importance of aptly selecting LLM APIs, even in the absence of budget constraints.

Next, we note that \systemFGPT{} enables smooth performance-cost trade-offs across all evaluated datasets. This offers flexible choices to LLM users and potentially helps LLM API providers save energy and reduce carbon emissions. In fact, \systemFGPT{} can simultaneously reduce costs and improve accuracy. For example, on the OVERRULING dataset, \systemFGPT{} achieves a 1\% accuracy gain while reducing costs by 73\% compared to the best LLM API, GPT-4. This is likely because \systemFGPT{} integrates knowledge from multiple LLMs. 

The example queries shown in Figure \ref{fig:AML:Tradeoffs} further aid in understanding why \systemFGPT{} can simultaneously improve performance and reduce costs. GPT-4 makes mistakes on some queries (e.g., the first example in part (a)), but some low-cost APIs provide correct predictions. \systemFGPT{} accurately identifies those queries and relies solely on the inexpensive APIs. For example, GPT-4 incorrectly infers no overruling from the legal statement "The time has come to reconcile and regularize our cases in this field," as shown in Figure \ref{fig:AML:Tradeoffs}(b). However, \systemFGPT{} accepts GPT-J's correct answer, avoiding the use of expensive LLMs and improving overall performance. Naturally, a single LLM API is not always correct; LLM cascade overcomes this by employing a chain of LLM APIs. For example, in the second example shown in Figure \ref{fig:AML:Tradeoffs}(a), \systemFGPT{} identifies that GPT-J's generation may not be reliable and turns to the second LLM in the chain, J1-L, to find the correct answer. Again, GPT-4 provides the wrong answer. \systemFGPT{} is not perfect, and there remains ample room for cost reduction. For example, in the third example in Figure \ref{fig:AML:Tradeoffs}(c), all LLM APIs in the chain give the same answer. However, \systemFGPT{} is unsure if the first LLMs are correct, resulting in the need to query all LLMs in the chain. Identifying how to avoid such cases remains an open problem.

\section{Discussions, Limitations and Future Prospects}\label{sec:AML:conclusion}
The substantial cost of employing LLMs in real-world scenarios presents a considerable barrier to their widespread usage. In this paper, we outline and discuss practical strategies for reducing the inference cost of using LLM APIs. We also developed \systemFGPT{} to illustrate one of the cost-saving strategies, LLM cascade.  
Our empirical findings show that \systemFGPT{} can reduce costs by up to 98\% while preserving the performance of cutting-edge LLMs.

\systemFGPT{} lays the groundwork for optimizing task performance with LLM APIs under budget constraints; however, it has some limitations. To train the LLM cascade strategy in \systemFGPT{}, we need some labeled examples. And in order for the cascade to work well, the training examples should be from the same or similar  distribution as the test examples. Moreover, learning the LLM cascade itself requires resources. We view this as an one-time upfront cost; this is beneficial when the final query dataset is larger than the data used to train the cascade. There are also other promising strategies for cost saving, such as speeding up attention computation itself, that we do not discuss here. Given the rapid development of LLM, this paper is not meant to be comprehensive or to provide a definitive solution. Our goal is to lay a foundation for this important research agenda and to demonstrate that even simple cascade can already achieve promising savings. 

There are also many related directions  for future exploration. While \systemFGPT{} concentrates on balancing performance and cost, real-world applications call for the evaluation of other critical factors, including latency, fairness, privacy, and environmental impact. Incorporating these elements into optimization methodologies while maintaining performance and cost-effectiveness is an important avenue for future research. Furthermore, utilizing LLMs in risk-critical applications necessitates the careful quantification of uncertainty in LLM-generated outputs. As the field progresses, addressing the environmental ramifications of training and deploying LLMs demands a joint effort from LLM users and API providers. The continuous evolution of LLMs and their applications will inevitably unveil new challenges and opportunities, fostering further research and development in this dynamic field.

\newpage
\bibliographystyle{alpha}
\bibliography{sample}

\end{document}

%% file: macro.tex
\usepackage{wrapfig}
\usepackage{bbm}
\usepackage{hyperref}       
\usepackage{url}            
\usepackage{booktabs}       
\usepackage{amsfonts}       
\usepackage{microtype}      
\usepackage{microtype}
\usepackage{graphicx}
\usepackage{subcaption}
\usepackage{booktabs} 
\usepackage{amsthm,amsmath,amssymb}
\usepackage{float,url,amsfonts,alltt}
\usepackage{mathtools,rotating}
\usepackage{ifpdf,fancyvrb}
\usepackage{enumitem}

\usepackage{microtype}
\usepackage{graphicx}
\usepackage{booktabs} 
\usepackage{hyperref}

\usepackage{graphicx,xspace,verbatim,comment}
\usepackage{hyperref,array,color,balance,multirow}
\usepackage{balance,float,url,amsfonts,alltt}
\usepackage{mathtools,rotating,amsmath,amssymb}
\usepackage{color,ifpdf,fancyvrb,array}
\usepackage{etoolbox,listings}
\usepackage{bigstrut,morefloats}

\usepackage{pbox}
\usepackage{algorithm}
\usepackage{algpseudocode}

\DeclarePairedDelimiterX{\inp}[2]{\langle}{\rangle}{#1, #2}

\newcommand{\eat}[1]{}

\newcommand{\E}{\mathbb{E}}

\numberwithin{equation}{section}

\usepackage{amsmath}
\usepackage{accents}
\newlength{\dhatheight}

\algrenewcommand\algorithmicrequire{\textbf{Input:}}
\algrenewcommand\algorithmicensure{\textbf{Output:}}

\newcommand{\systemFGPT}{{FrugalGPT}}

\newcommand{\prompt}{p}
\newcommand{\query}{q}
\newcommand{\answer}{a}
\newcommand{\cost}{c}